\documentclass[]{article}

\usepackage{lineno,hyperref}

\usepackage{amssymb}
\usepackage{latexsym}

\usepackage[margin=3cm]{geometry}
\usepackage{booktabs}
\usepackage{caption}
\usepackage{gensymb}
\usepackage{graphicx}
\usepackage{amssymb}
\usepackage{amsmath}
\usepackage{amsfonts}
\usepackage{breqn}
  \usepackage[caption=false,font=footnotesize]{subfig}

\usepackage{amsmath}
\usepackage{amssymb}
\usepackage{color}
\usepackage{breqn}
\usepackage{mathrsfs}
\usepackage{color,soul}

\usepackage{array}
\newcolumntype{L}[1]{>{\raggedright\let\newline\\\arraybackslash\hspace{0pt}}m{#1}}
\newcolumntype{C}[1]{>{\centering\let\newline\\\arraybackslash\hspace{0pt}}m{#1}}
\newcolumntype{R}[1]{>{\raggedleft\let\newline\\\arraybackslash\hspace{0pt}}m{#1}}

\usepackage{amsthm}

\begin{document}

		\title{Deep Supervised Hashing leveraging Quadratic Spherical Mutual Information for Content-based Image Retrieval}
		\author{Nikolaos Passalis and Anastasios Tefas\footnote{Dept. of Informatics, Aristotle University of Thessaloniki, Thessaloniki, 54124, Greece, email:   passalis@csd.auth.gr, tefas@aiia.csd.auth.gr }}	

\date{}	
		\maketitle
		
\begin{abstract}
Several deep supervised hashing techniques have been proposed to allow for efficiently querying large image databases. However, deep supervised image hashing techniques are developed, to a great extent, heuristically often leading to suboptimal results.  Contrary to this,  we propose an efficient deep supervised hashing algorithm that optimizes the learned codes using an information-theoretic measure, the Quadratic Mutual Information (QMI). The proposed method is adapted to the needs of large-scale hashing and information retrieval leading to a novel information-theoretic measure, the Quadratic Spherical Mutual Information (QSMI).  Apart from demonstrating the effectiveness of the proposed method under different scenarios and outperforming existing state-of-the-art image hashing techniques, this paper provides a structured way to model the process of information retrieval and develop novel methods adapted to the needs of each application.			
\end{abstract}

	\section{Introduction}

	The vast amount of data available nowadays, combined with the need to efficiently provide quick answers to users' queries led to the development of several \textit{hashing} techniques~\cite{zheng2014coupled, li2013hashing, kaushik2013efficient, yao2016semantic, xu2016learning, zhao2017spatial, ding2017large, ma2018global, pachori2018hashing}. Hashing provides a way to represent objects using compact codes, that allow for performing fast and efficient queries in large object databases. Early hashing methods, e.g., Locality Sensitive Hashing (LSH)~\cite{datar2004locality}, focused on extracting generic codes that could, in principle, describe every possible object and information need. However, it was later established that \textit{supervised hashing}, that learns hash codes that are tailored to the task at hand, can significantly improve the retrieval precision. In this way, it is possible to learn even smaller hashing codes, since the extracted code must only encode the information needs for which the users are actually interested in. However, note that the extracted hash codes must also encode part of the semantic relationships between the encoded objects, to allow for providing a meaningful ranking of the retrieved results.

	Many supervised and semi-supervised hashing methods have been proposed~\cite{Lai_2015_CVPR,xia2014supervised,zhang2015bit,zhao2015deep,zhu2016deep, yao2013semi, wang2015semi}. However, these methods were, to a great extent, heuristically developed, without a solid theory regarding the actual retrieval process. For example, many methods employ the pairwise distances between the images~\cite{Lai_2015_CVPR,xia2014supervised,zhu2016deep}, or are based on sampling \textit{triplets} that must satisfy specific relationships according to the given ground truth~\cite{zhang2015bit,zhao2015deep}, without a proper theoretical motivation for these choices. On the other hand, \textit{information-theoretic} measures, such as entropy and mutual information~\cite{principe2010information}, have been proven to provide robust solutions to many machine learning problems, e.g., classification~\cite{principe2010information}. However, very few steps towards using these measures for supervised hashing tasks have been made so far.

	In this paper, we provide a connection between an information-theoretic measure, the Mutual Information (MI)~\cite{principe2010information}, and the process of information retrieval. More specifically, we argue that mutual information can naturally model the process of information retrieval, providing a solid framework to develop retrieval-oriented supervised hashing techniques. Even though MI provides a sound theoretical formulation for the problem of information retrieval, applying it in real scenarios is usually intractable, since there is no efficient way to calculate the actual probability densities, that are involved in the calculation of MI. The great amount of data as well as their high dimensionality further complicate the practical application of such measures.

\begin{figure*}[t!]
		\begin{center}
			\includegraphics[width=0.99\linewidth]{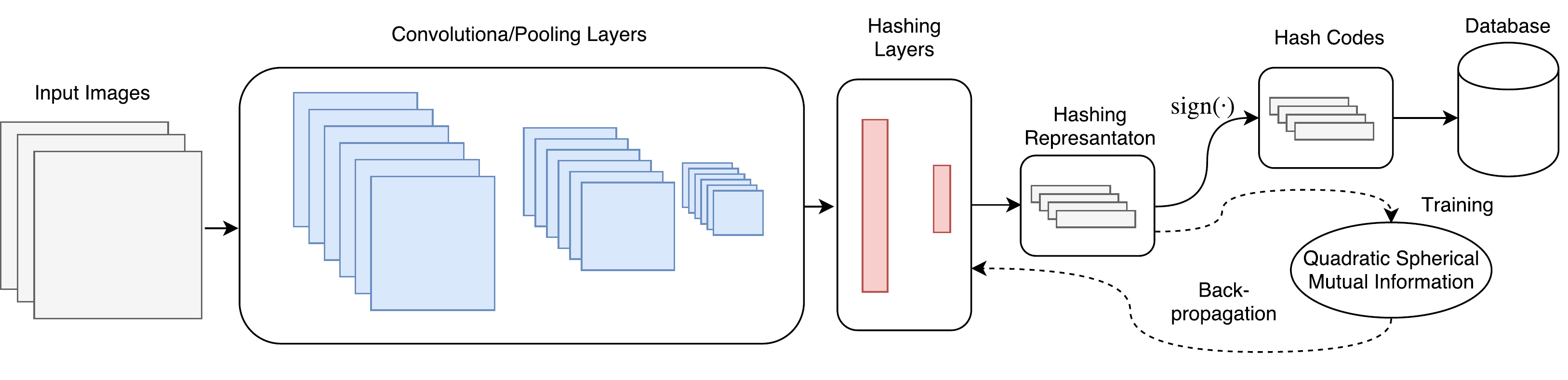}
		\end{center}
		\caption{Pipeline of the proposed method: A deep convolutional neural network (CNN) is used to extract a representation that can be used to directly obtain a binary hash code. The network is optimized using the proposed Quadratic Spherical Mutual Information loss that is adapted towards the needs of large-scale image hashing.}
		\label{fig:proposed}
	\end{figure*}

	The main contribution of this paper is the proposal of an efficient deep supervised hashing algorithm that optimizes the learned codes using a novel extension of an information-theoretic measure, the Quadratic Mutual Information (QMI)~\cite{torkkola2003feature}. The architecture of the proposed method is shown in Fig.~\ref{fig:proposed}.

	To derive a practical algorithm that can efficiently scale to large datasets:
	\begin{enumerate}
		\item We adapt QMI to the needs of supervised hashing by employing a similarity measure that is closer to the actual distance used for the retrieval process, i.e., the Hamming distance. This gives rise to the proposed \textit{Quadratic Spherical Mutual Information} (QSMI). It is also experimentally demonstrated that the proposed QSMI is more robust compared to the classical Gaussian-based Kernel Density Estimation used in QMI~\cite{torkkola2003feature}, while it does not require carefully tuning of any hyper-parameters.
		\item We propose using a more smooth optimization objective employing a novel square clamping approach. This allows for significantly improving the stability of the optimization, while reducing the risk of converging to bad local minima.
		\item We adapt the proposed approach to work in batch-based setting by employing a method that dynamically estimates the prior probabilities, as they are observed within each batch. In this way, the proposed method can efficiently scale to larger datasets.
		\item We demonstrate that the proposed method can be readily extended to efficiently handle different scenarios, e.g., retrieval of unseen classes~\cite{sablayrolles2017should}.
	\end{enumerate}
	
	The proposed method is extensively evaluated using three image datasets, including the two standard datasets used for evaluating supervised hashing methods, the CIFAR10~\cite{krizhevsky2009learning} and  NUS-WIDE~\cite{nus-wide-civr09} datasets, and it is demonstrated that it outperforms the existing state-of-the-art techniques. Following the suggestions of~\cite{sablayrolles2017should}, we also evaluate the proposed method in a different evaluation setup, where the learned hash codes are evaluated using unseen information needs. A PyTorch-based implementation of the proposed method is available at \url{https://github.com/passalis/qsmi}\footnote{The code will be available shortly after the review process.}, allowing any researcher to easily use the proposed method and readily reproduce the experimental results.

	The rest of the paper is structured as follows. The related work is discussed in Section~\ref{section:related}. The proposed method is presented in detail in Section~\ref{section:proposed}, while the experimental evaluation is provided in Section~\ref{section:evaluation}. Finally, Section~\ref{section:conclusions} concludes the paper.

\section{Related Work}
\label{section:related}

	The increasing interest for learning compact hash codes, together with the great learning capacity of recent deep learning models, led to the development of several {deep supervised hashing} techniques. Deep supervised hashing techniques involve: a) a deep neural network, that is used to extract a representation from the data, b) a supervised loss function, that is used to train the network, and c) a hashing mechanism, e.g., an appropriate non-linearity~\cite{li2017deep} or regularizer~\cite{zhu2016deep}, that ensures that the output of the network can be readily transformed into a hash code. Most of the proposed methods fall into one of the following two categories according to the loss function employed for learning the supervised codes: a) pairwise-based hashing methods~\cite{Lai_2015_CVPR,xia2014supervised,zhu2016deep,li2017deep,liu2016deep,jiang2017asymmetric,li2016feature,shenassymetric} and b) triplet-based hashing methods~\cite{zhang2015bit,zhao2015deep,wang2016deep}.

	Pairwise-based methods work by learning hash codes that minimize / maximize the pairwise distance / log-likelihood between similar / dissimilar pairs, e.g., Convolutional Neural Network (CNN)-based hashing~\cite{xia2014supervised}, network in network hashing~\cite{Lai_2015_CVPR}, deep hashing network~\cite{zhu2016deep}, deep pairwise-supervised hashing~\cite{li2016feature}, deep hashing network~\cite{zhu2016deep}, and  deep supervised discrete hashing~\cite{li2017deep}. More advanced pairwise methods employ margins that allow for learning more regularized representations, e.g., deep supervised hashing~\cite{liu2016deep}, use asymmetric hashing schemes, e.g., deep asymmetric pairwise hashing~\cite{shenassymetric}, and asymmetric deep supervised hashing \cite{jiang2017asymmetric}, or use more advanced techniques to obtain the binary codes, e.g., hashing by continuation~\cite{Cao_2017_ICCV}. Even though these techniques have been applied with great success, they were largely developed heuristically. On the other hand, the proposed method works by maximizing the mutual information between the learned hash codes and the ground truth.

	Triplet-based methods work by sampling an anchor point along with a positive and a negative example~\cite{zhang2015bit,zhao2015deep,wang2016deep,deng2018triplet}. Then, they learn codes that increase the similarity between the anchor and the positive example, while reducing the similarity between the anchor and the negative example. However, triplet-based methods are significantly more computationally expensive than pairwise-based methods, requiring a huge number of triplets to be generated (many of which convey no information, since they are already satisfied by the code learned by the network), limiting their practical application.  Also note that many non-deep supervised hashing methods have also been  proposed, e.g., \cite{kang2016column,lin2014fast,liu2012supervised}, but an extensive review of them is out of the scope of this paper. The interested reader is referred to~\cite{wang2017survey} for an extensive literature review on hashing.

	The use of MI has been also investigated to aid various aspects of the retrieval process. In~\cite{almasri2016comparison,hu2006improving} MI is employed to provide relevance feedback, while in~\cite{Cakir_2017_ICCV} MI is used to provide  updates for online hashing. More specifically, the Shannon's definition for MI is used in~\cite{Cakir_2017_ICCV}, leading to employing a Monte Carlo sampling scheme to approximate the MI, together with a differentiable histogram binning technique. Our approach is vastly different, since instead of approximating the MI through random sampling, we analytically derive computationally tractable solutions for calculating MI through a QMI formulation. Furthermore, we also adapt MI to the actual needs of hashing by employing a spherical formulation that is closer to the Hamming distance. Note that other information-theoretic criteria, such as entropy~\cite{passalis2016entropy, passalis2017learning, passalis2018learning}, have been also employed to optimize various representations towards information retrieval.

	To the best of our knowledge, this is the first work that employs a quadratic spherical mutual information loss fully adapted to the needs of deep supervised hashing. Apart from deriving a practical algorithm and demonstrating its ability to outperform existing state-of-the-art methods, the proposed method provides a complete framework that can be used to model the process of information retrieval. This formulation is fully differentiable allowing for the end-to-end optimization of deep neural networks for any retrieval-related task, ranging from learning retrieval-oriented representations and compact hash codes to fine-tuning the extracted representations using relevance feedback.

\section{Proposed Method}
\label{section:proposed}

	The proposed method is presented in detail in this Section. First, the links between mutual information and information retrieval are provided. Then, the quadratic mutual information is introduced, the proposed quadratic spherical mutual information is derived and several aspects of the proposed method are discussed. 
	
	\subsection{Information Retrieval and Mutual Information} 
	
	Let $\mathcal{Y}=\{\mathbf{y}_1, \mathbf{y}_2, ..., \mathbf{y}_N\}$ be a collection of $N$ images, where $\mathbf{y}_i \in \mathbb{R}^n$ is the representation of the $i$-th image extracted using an appropriate feature extractor, e.g., a deep neural network. Each image $\mathbf{y}_i$ fulfills a set of information needs. For example, an image that depicts a ``red car near a beach'' fulfills at least the following information needs: ``car'', ``red car'', ``beach'', ``car near beach''. Note that the information needs that an image actually fulfills depend on both its content and the needs of the users, since, depending on the actual application, the interests of the users are usually focused on a specific area. For example, an image of a man entering a bank represents different information needs for a forensics database used by the police to identify suspects and for a generic web search engine. The problem of information retrieval can be then defined as follows: \textit{Given an information need ${q}$ retrieve the images of the collection $\mathcal{Y}$ that fulfill this information need and rank them according to their relevance to the given information need}. This work focuses on \textit{content-based} information retrieval~\cite{lew2006content}, where the information need $q$ is expressed through a \textit{query image} $\mathbf{q} \in \mathbb{R}^n$, that is usually not part of the collection $\mathcal{Y}$.

	To be able to measure how well an information retrieval system works, a ground truth set that contains a set of information needs and the corresponding images that fulfill these information needs is usually employed. Let $M$ be the number of information needs $\mathcal{Q}=\{q_1, q_2, ..., q_M\}$. Then, for each information need $q_i$, a set of images $\mathcal{Q}_{i} = \{\mathbf{y}^{(i)}_1, \mathbf{y}^{(i)}_2, ..., \mathbf{y}^{(i)}_{N_i}\}$, where $\mathbf{y}^{(i)}_j \in \mathbb{R}^n$ is the representation of the $j$-th image that fulfills the $i$-th information need, is given. Note that $|\mathcal{Q}_{i}| = N_i$. Since all these images fulfill the same information need, they can be all used as queries to express this information need. However, there are also other images, which are usually not known beforehand, that also express the same information need and they can be also used to query the database. The distribution of the images that fulfill the $i$-th information need can be modeled using the \textit{conditional probability density} function $p(\mathbf{y}|q_i)$.

	Let $\mathscr{Y}$ be a random vector that represents the images and $\mathscr{Q}$ be a random variable that represents the information needs. The Shannon's entropy of the information needs, that expresses the uncertainty regarding the information need that a randomly sampled image fulfills, is defined as~\cite{principe2010information}:
	\begin{equation}
	H(\mathscr{Q}) = - \sum_{q} P(q) log(P(q)),
	\end{equation}
	where $P(q)$ is the prior probability of the information need $q$, i.e., the probability that a random image of the collection fulfills the information need $q$. Note that above definition implicitly assumes that the information needs are mutually exclusive, i.e.,  $\sum_q P(q) = 1$, or equivalently, that each image satisfies only one information need. This is without loss of generality, since it is straightforward to extend this definition to the general case, where each image can satisfy multiple information needs, simply by measuring the entropy of each information need separately: $
	H(\mathscr{Q}) = - \sum_{q} \Big( P(q) log(P(q)) + (1-P(q)) log(1-P(q))\Big).$

	To simplify the presentation of the proposed method, we assume that the information needs are mutually exclusive. Nonetheless, the proposed approach can be still used with minimal modifications,  as we also experimentally demonstrate in Section~\ref{section:evaluation}, even when this assumption does not hold.
	
	When the query vector is known, then the uncertainty of the information need that it fulfills can be expressed by the conditional entropy:
	\begin{equation}
	H(\mathscr{Q}|\mathscr{Y}) = -\int_{\mathbf{y}}p(\mathbf{y})\left(\sum_{q}^{}p(q|\mathbf{y}) log(p(q|\mathbf{y}))\right)d\mathbf{y}.
	\end{equation}

	Mutual information is defined as the amount by which the uncertainty for the information needs is reduced after observing the query vector:
	\begin{equation}
	\label{eq:mi}
	I(\mathscr{Q}, \mathscr{Y}) = H(\mathscr{Q})  - H(\mathscr{Q}|\mathscr{Y}) = \sum_{q} \int_{\mathbf{y}} p(q, \mathbf{y}) log\left(\frac{p(q, \mathbf{y})}{P(q)p(\mathbf{y})}\right)d\mathbf{y}.
	\end{equation}
	
	It is easy to see that MI can be interpreted as the Kullback-Leibler divergence between $p(q, \mathbf{y})$ and $P(q)p(\mathbf{y})$. It is desired to maximize the MI between the representation of the images $\mathscr{Y}$ and the information needs $\mathscr{Q}$, since this ensures that the uncertainty regarding the information need, that a query image expresses, is minimized. Also, note that MI models the intrinsic uncertainty regarding the query vectors, since it employs the conditional probability density between the information needs and the images, instead of just a limited collection of images. 
	
	On the other hand, it is usually intractable to directly calculate the required probability density $p(\mathbf{y}|q_i)$ and the corresponding integral in~(\ref{eq:mi}), limiting the practical applications of MI. However, as it is demonstrated later, it is possible to efficiently estimate the aforementioned probability density and derive a practical algorithm that maximizes the MI between a representation and a set of information needs.

	\subsection{Quadratic Mutual Information}
	
	When the aim is not to calculate the exact value of MI, but to optimize a distribution that maximizes the MI, then a quadric divergence metric, instead of the Kullback-Leibler divergence, can be used. In this way, the \textit{Quadratic Mutual Information} (QMI) is defined as~\cite{torkkola2003feature}:
	\begin{equation}
	\label{eq:qmi}
	I_T (\mathscr{Q}, \mathscr{Y}) = \sum_{q} \int_{\mathbf{y}} \left(p(q, \mathbf{y}) - P(q)p(\mathbf{y}) \right)^2 d\mathbf{y}.
	\end{equation}

	By expanding~(\ref{eq:qmi}), QMI can be expressed as the sum of three \textit{information potentials} as $I_T (\mathscr{Q}, \mathscr{Y}) = V_{IN} (\mathscr{Q}, \mathscr{Y}) + V_{ALL} (\mathscr{Q}, \mathscr{Y}) - 2V_{BTW}  (\mathscr{Q}, \mathscr{Y})$, where: $V_{IN} (\mathscr{Q}, \mathscr{Y}) = \sum_{q} \int_{\mathbf{y}} p(q, \mathbf{y})^2 d\mathbf{y}$, $
	V_{ALL} (\mathscr{Q}, \mathscr{Y}) = \sum_{q} \int_{\mathbf{y}} P(q)^2p(\mathbf{y})^2 d\mathbf{y}$, and $V_{BTW} (\mathscr{Q}, \mathscr{Y}) = \sum_{q} \int_{\mathbf{y}} p(q, \mathbf{y})P(q)p(\mathbf{y}) d\mathbf{y}$.

	To calculate these quantities, the probability $P(q)$ and the densities $p(\mathbf{y})$ and $p(q, \mathbf{y})$ must be estimated. The prior probabilities depend only on the distribution of the information needs in the collection of images. Therefore, for the $i$-th information need: $P(q_i) = \frac{N_i}{N}$, where $N_i$ is the number of images that fulfill the $i$-th information need. The conditional density of the images that fulfill the $i$-th information need can be estimated using the Parzen window estimation method~\cite{parzen1962estimation}:
	\begin{equation}
	p(\mathbf{y}|q_i) = \frac{1}{N_i} \sum_{j=1}^{N_i} K(\mathbf{y}-\mathbf{y}^{(i)}_j; \sigma^2),
	\end{equation}
	where $K(\mathbf{y}; \sigma^2)$ is a Gaussian kernel (in an $n$-dimensional space) with width $\sigma$ defined as:
	
	\begin{equation}
	\label{eq:gaussina}
	K(\mathbf{y}; \sigma) = \frac{1}{{(2\pi)^{n/2}} \sqrt{\sigma}} exp(- \frac{\mathbf{y}^T\mathbf{y}}{2\sigma}).
	\end{equation}
	
	Then, the joint probability density can be estimated as:
	\begin{equation}
	p(q_i, \mathbf{y})= p(\mathbf{y}|q_i)p(q_i) = \frac{1}{N} \sum_{j=1}^{N_i} K(\mathbf{y}-\mathbf{y}^{(i)}_j; \sigma^2),
	\end{equation}
	while the density of all the images as:
	\begin{equation}
	p(\mathbf{y}) = \frac{1}{N} \sum_{j=1}^{N} K(\mathbf{y}-\mathbf{y}^{(i)}_j; \sigma^2).
	\end{equation}
	
	By substituting these estimations into the definitions of the information potentials, the following quantities are obtained:
	\begin{equation}
	\label{eq:v1}
	V_{IN} (\mathscr{Q}, \mathscr{Y}) = \frac{1}{N^2} \sum_{k=1}^M \sum_{i=1}^{N_k} \sum_{j=1}^{N_k} K(\mathbf{y}^{(k)}_i - \mathbf{y}^{(k)}_j, 2\sigma^2),
	\end{equation}
	\begin{equation}
	\label{eq:v2}
	V_{ALL} (\mathscr{Q}, \mathscr{Y}) = \frac{1}{N^2} \left( \sum_{k=1}^M \left(\frac{N_k}{N}\right)^2\right)  \sum_{i=1}^{N} \sum_{j=1}^{N} K(\mathbf{y}_i - \mathbf{y}_j, 2\sigma^2),
	\end{equation}
	and
	\begin{equation}
	\label{eq:v3}
	V_{BTW} (\mathscr{Q}, \mathscr{Y}) = \frac{1}{N^2} \sum_{k=1}^{M} \left( \left(\frac{N_k}{N}\right)  \sum_{i=1}^{N_k} \sum_{j=1}^{N}  K(\mathbf{y}^{(k)}_i - \mathbf{y}_j, 2\sigma^2) \right),
	\end{equation}
	where the following property regarding the convolution between two Gaussian kernels was used: $\int_{\mathbf{y}} K(\mathbf{y}-\mathbf{y}_i; \sigma^2)  K(\mathbf{y}-\mathbf{y}_j; \sigma^2)d\mathbf{y} = K(\mathbf{y}_i-\mathbf{y}_j, 2\sigma^2)$. The information potential $V_{IN}$ expresses the interactions between the images that fulfill the same information need, the information potential $V_{ALL}$ the interactions between all the images of the collection, while the potential $V_{BTW}$ models the interactions of the images that fulfill a specific information need against all the other images. Therefore, the QMI formulation allows for the efficient calculation of MI, since the MI is expressed as a weighted sum over the pairwise interactions of the images of the collection.

	Using Parzen window estimation with a Gaussian kernel for estimating the probability density leads to the implicit assumption that the similarity between two images is expressed through their Euclidean distance. Thus, the images that fulfill an information need expressed by a query vector $\mathbf{q}$ can be retrieved simply using nearest-neighbor search.

	\subsection{Quadratic Spherical Mutual Information Optimization}

	Even though QMI allows for more efficient optimization of distributions, it suffers from several limitations: a) QMI involves the calculation of the pairwise similarity matrix between all the images of a collection. This quickly becomes intractable as the size of the collection increases. b) Selecting the appropriate width for the Gaussian kernels is not always straightforward, as a non-optimal choice can distort the feature space and slow down the optimization. c) The discrepancy between the distance metric used for QMI (Euclidean distance) and the distance used for the actual retrieval of the hashed images (Hamming distance) can negatively affect the retrieval accuracy. Finally, d) it was experimentally observed that directly optimizing the QMI is prune to bad local minima, due to the linear behavior of the loss function that fails to distinguish between the pairs of images that cause high error and those which have a smaller overall effect on the learned representation (more details are given later in this Section).

	To overcome the limitations (b) and (c), we propose the \textit{Quadratic Spherical Mutual Information} (QSMI). The proposed QSMI method replaces the Gaussian kernel in (\ref{eq:gaussina}), used for calculating the similarity between two images in the information potentials in~(\ref{eq:v1}),~(\ref{eq:v2}), and (\ref{eq:v3}), with the cosine similarity:
	\begin{equation}
	\label{eq:cosine}
	S_{cos}(\mathbf{y}_1, \mathbf{y_2}) = \frac{1}{2} \left(\frac{\mathbf{y}^T_1 \mathbf{y}_2}{\Vert \mathbf{y}_1\Vert_2 \Vert\mathbf{y}_2\Vert_2} + 1 \right),
	\end{equation}
	where $\Vert \cdot \Vert_2$ is the $l^2$ norm of a vector.
	In this way, we maintain the computationally efficient QMI formulation and avoid the need for manually tuning the width parameter of the Gaussian kernel, while adopting a formulation that is closer to the Hamming distance that is actually used for the retrieval process. Indeed, the cosine similarity can be interpreted as a normalized Hamming-based similarity measure~\cite{liu2012supervised}. To understand this, consider that the Hamming ``similarity'' (number of bits that are equal between two binary vectors) can be calculated as $\mathbf{y}_1^T\mathbf{y}_2$, for two binary $n$-dimensional vectors $\mathbf{y}_1, \mathbf{y}_2 \in \{0, 1\}^n$. Therefore, when binary vectors are used, the cosine similarity can be interpreted as a bounded normalized version of the Hamming similarity.
	
	Therefore, QSMI is defined as:
	\begin{equation}
		I^{cos}_T (\mathscr{Q}, \mathscr{Y}) = V^{cos}_{IN} (\mathscr{Q}, \mathscr{Y}) + V^{cos}_{ALL} (\mathscr{Q}, \mathscr{Y}) - 2V^{cos}_{BTW} (\mathscr{Q}, \mathscr{Y}),
	\end{equation}
	where
	\begin{equation}
	V^{cos}_{IN} (\mathscr{Q}, \mathscr{Y}) = \frac{1}{N^2} \sum_{k=1}^M \sum_{i=1}^{N_k} \sum_{j=1}^{N_k} S_{cos}(\mathbf{y}^{(k)}_i, \mathbf{y}^{(k)}_j),
	\end{equation}
	\begin{equation}
	V^{cos}_{ALL} (\mathscr{Q}, \mathscr{Y}) = \frac{1}{N^2} \left( \sum_{k=1}^M \left(\frac{N_k}{N}\right)^2\right)  \sum_{i=1}^{N} \sum_{j=1}^{N} S_{cos}(\mathbf{y}_i, \mathbf{y}_j),
	\end{equation}
	and
	\begin{equation}
	V^{cos}_{BTW} (\mathscr{Q}, \mathscr{Y}) = \frac{1}{N^2} \sum_{k=1}^{M} \left( \left(\frac{N_k}{N}\right)  \sum_{i=1}^{N_k} \sum_{j=1}^{N}  S_{cos}(\mathbf{y}^{(k)}_i, \mathbf{y}_j) \right).
	\end{equation}

	Note that when the information needs are equiprobable, i.e., $P(q) = \frac{1}{M}$, then QSMI can be simplified as $I^{cos}_T (\mathscr{Q}, \mathscr{Y}) = V^{cos}_{IN} (\mathscr{Q}, \mathscr{Y})  - V^{cos}_{BTW} (\mathscr{Q}, \mathscr{Y})$. Therefore, when this assumption holds, QSMI can be easily implemented just by defining the similarity matrix $\mathbf{S} \in \mathbb{R}^{N\times N}$, where $[\mathbf{S}]_{ij} = S_{cos}(\mathbf{y}_i, \mathbf{y}_j)$ and the notation $[\mathbf{S}]_{ij}$ is used to refer to the $i$-th row and $j$-th column of matrix $\mathbf{S}$. Then, QSMI can be calculated as:
	\begin{equation}
	\label{eq:qsmi-original}
	I^{cos}_T = \frac{1}{N^2} \mathbf{1}^T_N \left(\mathbf{\Delta}\odot\mathbf{S} - \frac{1}{M}\mathbf{S}\right) \mathbf{1}_N,
	\end{equation}
	where the indicator matrix is defined as:
	\begin{equation}
	[\mathbf{\Delta}]_{ij} = 
	\begin{cases}
	1, & \text{if the $i$-th and the $j$-th documents are similar}\\
	0, & \text{otherwise}.
	\end{cases}
	\end{equation}

	The notation $\mathbf{1}_N \in \mathbb{R}^N$ is used to refer an $N$-dimensional vector of 1s, while the operator $\odot$ denotes the Hadamard product between two matrices. Please refer to the Appendix~A for a more detailed derivation. This formulation also allows for directly handling information needs that are not mutually exclusive.  In that case, the values of the indicator matrix are appropriately set to 1, if two images share at least one information need.

	Instead of directly optimizing the QSMI, we propose using a ``square clamp'' around the similarity matrix $\mathbf{S}$, smoothing the optimization surface. Therefore, given that the values of $\mathbf{S}$ range in the unit interval, the loss function is re-derived as:
	
	\begin{equation}
		\label{eq:qsmi-obj}
		\mathcal{L}_{QSMI} = \frac{1}{N^2} \mathbf{1}^T_N \left(\mathbf{\Delta}\odot (\mathbf{S}-1)\odot(\mathbf{S}-1) + \frac{1}{M}(\mathbf{S}\odot\mathbf{S})\right) \mathbf{1}_N.
	\end{equation}

	As shown in Figure~\ref{fig:smi-sim} this formulation penalizes  the pairs with larger error more heavily than those with smaller error, allowing for discovering more robust solutions. This modification effectively addresses the limitation (d), as we also experimentally demonstrate in the ablation study given in Section~\ref{section:evaluation}.

	To allow for scaling to larger datasets, batch-based optimization is used. That allows for reducing the complexity of QMI from $O(N^2)$, which is intractable for larger datasets, to just $O(N_B^2)$, where $N_B$ is the used batch size that typically ranges from 64 to 256. However, this implies that each batch will contain images only from a subsample of the available information needs. This in turn means that the observed in-batch prior probability $P(q)$ will not match the collection-level prior, leading to underestimating the influence of the potential $V_{ALL}$ to the optimization. To account for this discrepancy, we propose a simple heuristic to estimate the in-batch prior, i.e., the value of $M$ in~(\ref{eq:qsmi-obj}): $M$ is estimated as $M = N^2_B / (\mathbf{1}^T_{N_B} \mathbf{\Delta} \mathbf{1}_{N_B})$, where $N_B$ is the batch size. To understand the motivation behind this, consider that if the whole collection was used for the optimization, then the number of 1s in $\mathbf{\Delta}$ would be:  $M (\frac{N}{M})^2 = \mathbf{1}^T_{N} \mathbf{\Delta} \mathbf{1}_{N}$. Solving this equation for $M$ yields the value used for approximating $M$. Note that the value of $M$ is not constant and depends on the distribution of the samples in each batch. It was experimentally verified that this approach indeed improves the performance of the proposed method over using a constant value for $M$.

	\subsection{Deep Supervised Hashing using QSMI}
	
	The proposed QSMI is used to train a deep neural network to extract short binary hash codes, as shown in Fig.~\ref{fig:proposed}. Let $\mathbf{x}$ be the raw representation of an image (e.g., the pixels of an image) and let $\mathbf{y} = f_{\mathbf{W}}(\mathbf{x}) \in \mathbb{R}^n$ be the output of a neural network $f_{\mathbf{W}}(\cdot)$, where $\mathbf{W}$ denotes the matrix of the parameters of the network and $n$ is the length of the hash code. Apart from learning a representation that minimizes the $J_{QSMI}$ loss, the network must generate an output that can be easily translated into a binary hash code. Several techniques have been proposed to this end, e.g., using the \textit{tanh} function ~\cite{wang2017survey}. In this work, the output of the network is required to be close to two possible values, either 1 or -1. Therefore, the used hashing regularizer is defined, following the recent deep supervised hashing approaches~\cite{liu2016deep}, as:
	\begin{equation}
	\label{eq:hashing}
	\mathcal{L}_{hash} = \sum_{i=1}^N \Vert |\mathbf{y}_i| - \mathbf{1}_n \Vert_1,
	\end{equation}
	where $|\cdot|$ denotes the absolute value operator and $ ||\cdot||_1$ denotes the $l^{1}$ norm. The final loss function is defined as:
	\begin{equation}
	\mathcal{L} = \mathcal{L}_{QSMI} + \alpha \mathcal{L}_{hash},
	\end{equation}
	where $\alpha$ is the weight of the hashing regularizer. The network $f_{\mathbf{W}}(\cdot)$ can be then trained using gradient descent, i.e., $\Delta \mathbf{W} = - \eta\frac{\partial J}{\partial \mathbf{W}}$, where $\eta$ is the used learning rate. Please refer to Appendix~A on details regarding the derivation of $\frac{\partial J}{\partial \mathbf{W}}$. After training the network, the hash codes can be readily obtained using the $sign(\mathbf{y})$ function.

	Even though learning highly discriminative hash codes is desired for retrieving data that belong to the training domain, it can negatively affect the retrieval for previously unseen information needs~\cite{sablayrolles2017should}. The proposed method can be also easily modified to optimize the hash codes toward other \textit{unsupervised} information needs. Even though any source of information can be used, in this work the information needs are discovered by clustering the training data. This allows for discovering information needs dictated by the structure of the data. Let $\mathcal{L}_{QSMI+U}$ denote the loss induced by applying the QMSI loss function on these information needs. Then, the final loss function for this semi-supervised variant is defined as: $\mathcal{L} = \mathcal{L}_{QSMI} + \alpha \mathcal{L}_{hash} + \beta \mathcal{L}_{QSMI+U}$. This variant is used for the experiments conducted in Section~\ref{section:eval-unseen}.

	\section{Experimental Evaluation}
	
	\label{section:evaluation}

	The proposed method is extensively evaluated in this Section, using both an ablation study and comparing it to other state-of-the-art methods. First, the used datasets and the employed evaluation setup are briefly described. The hyper-parameters and the network architectures used for the evaluation are provided in Appendix~B. Finally, an ablation study is provided and the proposed method is evaluated using three different datasets.

	\subsection{Datasets and Evaluation Metrics}

	Three image datasets are used to evaluate the proposed method in this paper: The Fashion MNIST dataset, the CIFAR10 dataset and the NUS-WIDE dataset. All images were preprocessed to have zero mean and unit variance, according to the statistics of the dataset used for the training.

	The Fashion MNIST dataset is composed of 60,000 training images and 10,000 test images~\cite{xiao2017fashion}. The size of each image is $28 \times 28$ pixels (gray-scale images) and there is a total of 10 different classes (each one expresses a different information need). The whole training set was used to train the networks and build the database, while the test set was used to query the database and evaluate the performance of the methods.

	The CIFAR10 dataset is composed of 50,000 training images and 10,000 test images~\cite{krizhevsky2009learning}. The size of each image is $32 \times 32$ pixels (color images) and there is a total of 10 different classes (information needs). Again, the whole training set was used to train the networks and build the database, while the test set was used to query the database and evaluate the performance of the methods.

	The NUS-WIDE is a large-scale dataset that contains 269,648 images that belong to 81 different concepts~\cite{nus-wide-civr09}. The images were resized to $224\times224$ pixels before feeding them to the network. Following~\cite{li2016feature}, only images that belong to the 21 most frequent concepts, i.e., 195,834 images, were used for training/evaluating the methods.  Each image might belong to multiple different concepts, i.e., the information needs are not mutually exclusive. For evaluating the methods, two images were considered relevant if they share at least one common concept, which is the standard protocol used for this dataset~\cite{li2016feature}. Similarly to the other two datasets, the whole training set (193,734 randomly sampled images) was used to train the networks and build the database, while 2,100 randomly sampled queries (100 from each category) were employed to evaluate the methods.

	To evaluate the proposed methods, the following four metrics were used: precision, recall, mean average precision (mAP), and precision within hamming radius of 2. Nearest neighbor search using the Hamming distance was used to retrieve the relevant documents \cite{manning2008introduction}. Following ~\cite{manning2008introduction}, precision is defined as $Pr(q, k) = \frac{rel(q, k)}{k}$, where $k$ is the number of retrieved objects and $rel(q, k)$ is the number of retrieved objects that fulfill the same information need as the query $q$, while recall is defined as  $Rec(q, k) = \frac{rel(q, k)}{ntotal(q)}$, where $ntotal(q)$ is the total number of database objects that fulfill the same information as as~$q$. The average precision (AP) is calculated at eleven equally spaced recall points (0, 0.1, ..., 0.9, 1) and the mean average precision is calculated as the mean of the APs for all queries.  The precision withing hamming radius of 2 is defined as: $Pr_{H2}(q) = \frac{rel_{H2}(q)}{total_{H2}(q)}$, where  $rel_{H2}(q)$ is the number of relevant documents withing hamming distance 2 from the query, while $total_{H2}(q)$ is the total number of documents withing hamming distance 2 from the query.

	\subsection{Ablation Study}

	\begin{figure*}[]
		\begin{center}
			\captionsetup[subfigure]{width=0.32\linewidth}
			
			\subfloat[][Comparing the behavior of different loss functions for various errors (the original QMI loss is shifted by 1 to allow for easily comparing the plots)]{\includegraphics[width=0.32\linewidth]{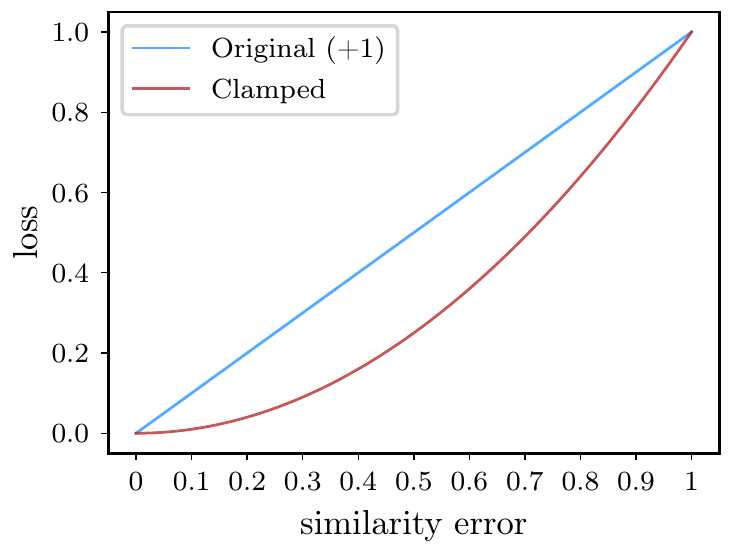}\label{fig:smi-sim} \hspace{5px}}
			\subfloat[][The proposed Clamped QSMI loss and the QSMI and QMI measures during the optimization (the plots have been normalized to the unit interval)]{\includegraphics[width=0.32\linewidth]{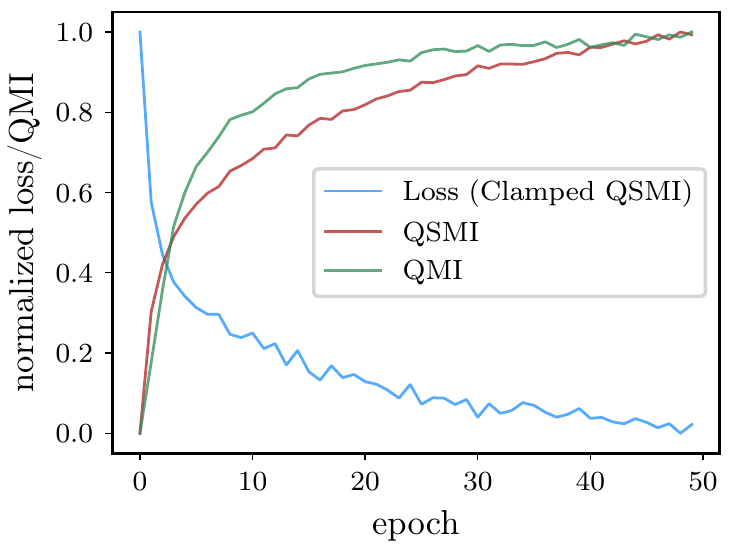}\label{fig:qsmi-loss} \hspace{5px}}
			\subfloat[][Effect of the regularization parameter $\alpha$ on the mAP and precision within hamming radius 2 ]{\includegraphics[width=0.32\linewidth]{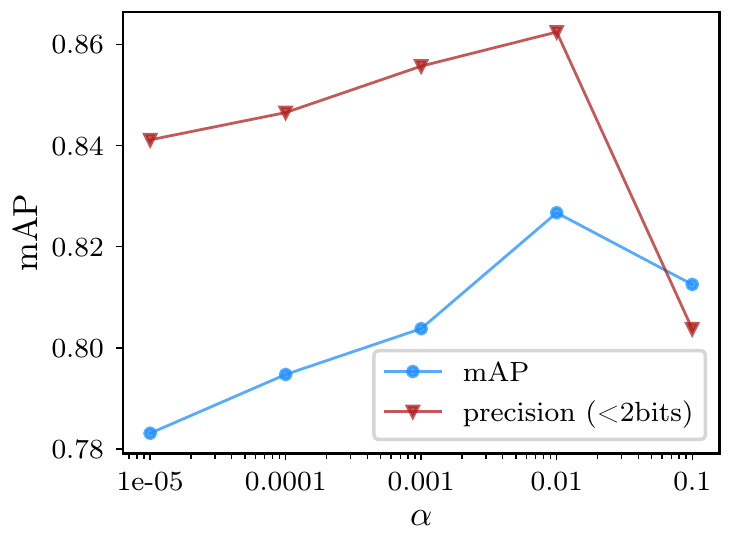}\label{fig:alpha}}
		\end{center}
		\caption{Ablation Study: Studying the differences between the original loss and the proposed hamming loss (Fig.~\ref{fig:smi-sim}), the effect of the optimization on the QMI and QSMI measures (Fig.~\ref{fig:qsmi-loss}), and the effect of the regularization parameter $\alpha$ (Fig.~\ref{fig:alpha})}
	\end{figure*}

	\begin{table}
		\caption{Ablation study using the Fashion MNIST dataset (the mAP is reported)}
		\label{table:ablation}
		\begin{center}
			\begin{tabular}{c|c||cc}
				\textbf{Clamped} & \textbf{Spherical} & \textbf{mAP} & \textbf{precision ($\mathbf{<2bits}$)} \\
				\hline
				No & No & ${0.727 \pm  0.008}$ & ${0.674 \pm 0.021}$ \\
				Yes & No & ${0.816 \pm  0.009}$ & ${0.864 \pm 0.010}$ \\
				Yes & Yes & $\mathbf{0.861 \pm  0.004}$ & $\mathbf{0.876 \pm 0.004}$ \\
			\end{tabular}
		\end{center}
	\end{table}

	A smaller dataset, the Fashion MNIST dataset~\cite{xiao2017fashion}, was used to perform an ablation study. The effect of various design choices, i.e., using or not the proposed clamped loss and spherical formulation, is evaluated in Table~\ref{table:ablation}. The mean Average Precision (mAP) is averaged over 5 runs, while the code length was set to 48 bits for these experiments. Several conclusions can be drawn from the results reported in Table~\ref{table:ablation}. First, employing the proposed clamped loss, instead of directly optimizing the QMI ($\sigma=10$), improves the hashing precision, confirming our hypothesis regarding the benefits of using the proposed clamped loss (as also described in the previous Section and shown in Fig.~\ref{fig:smi-sim}). This is also confirmed in the learning curve shown in Fig.~\ref{fig:qsmi-loss}, where both the proposed clamped loss and MI are monitored during the optimization. Optimizing the proposed clamped loss is directly correlated with the QMI and the proposed QSMI, both of which steadily increase during the 50 training epochs. When the spherical formulation is used (QSMI method), then the mAP further increase to 86.1\% from 72.7\% (standard QMI formulation).  The effect of the regularization parameter $\alpha$ is also examined in Fig.~\ref{fig:alpha}. The best performance is obtained for $\alpha=0.01$.

\begin{figure*}
		\begin{center}
						\captionsetup[subfigure]{width=0.32\linewidth}
			\subfloat[][Fashion MNIST]{\includegraphics[width=0.33\linewidth]{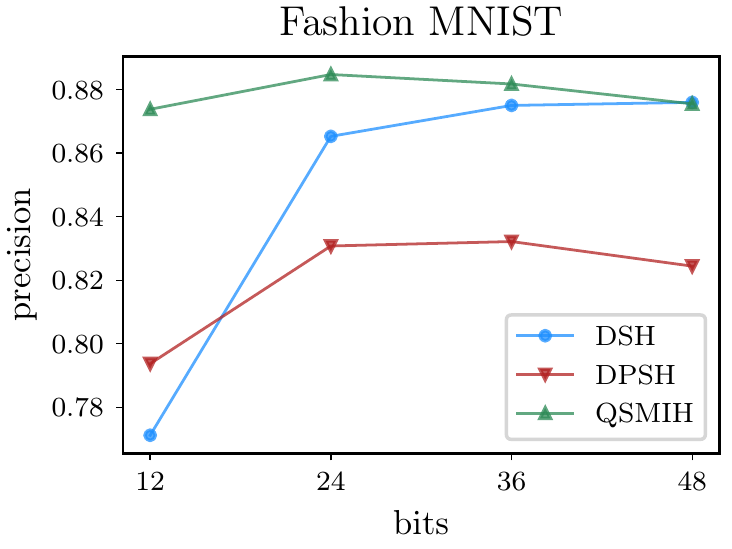}\label{fig:hamming-fashion}\hspace{5px}} 
			\subfloat[][CIFAR10]{\includegraphics[width=0.33\linewidth]{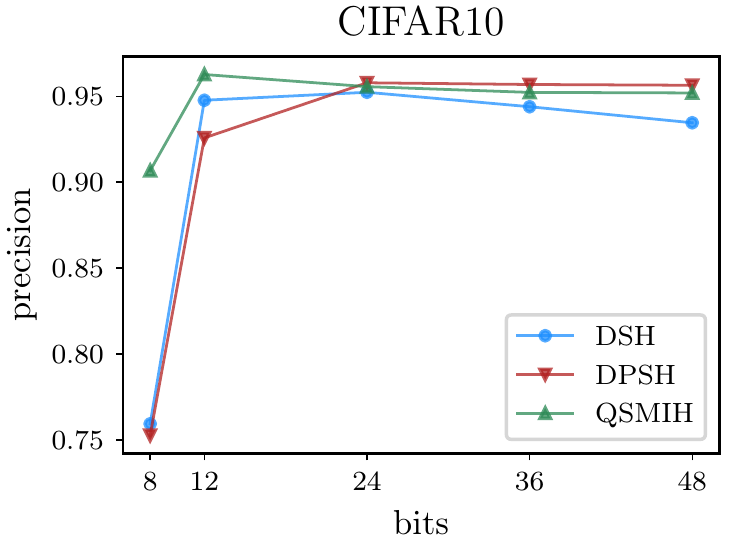}\label{fig:hamming-cifar}  \hspace{5px}}
			\subfloat[][NUS-WIDE]{\includegraphics[width=0.33\linewidth]{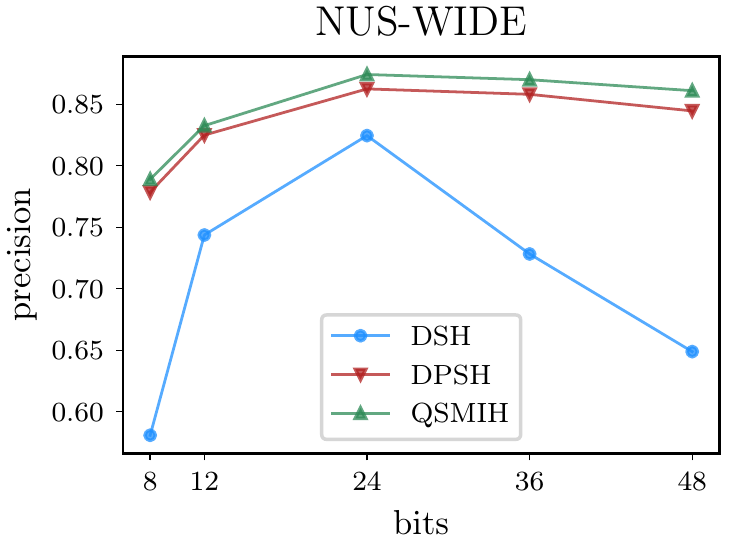}\label{fig:hamming-nuswide}}
			\hspace{1em}
		\end{center}
		\caption{Precision within hamming radius 2 for three different datasets}
	\end{figure*}

	The proposed method was compared to two other state-of-the-art techniques, the Deep Supervised Hashing (DSH) method~\cite{liu2016deep} and the Deep Pairwise Supervised Hashing (DPSH) method~\cite{li2016feature}. We carefully implemented these methods in a batch-based setting and we tuned their hyper-parameters to obtain the best performance (please refer to Appendix~B). The evaluation results are shown in Table~\ref{table:fashion}. The proposed method is abbreviated as ``QSMIH'' and significantly outperforms the other two state-of-the-art pairwise hashing techniques, highlighting the importance of using theoretically-sound objectives for learning deep supervised hash codes. Recall that a deep CNN, that was trained from scratch, was employed for the conducted experiments. The precision within Hamming radius of 2 is also shown in Figure~\ref{fig:hamming-fashion}. Again, the proposed method outperforms all the other methods for all the evaluated hash code lengths.

	\begin{table}
		\caption{Fashion MNIST Evaluation (the mAP for different hash code lengths is reported)}
		\label{table:fashion}

		\begin{center}
			\begin{tabular}{@{\hskip3pt}l@{\hskip3pt}|@{\hskip3pt}c@{\hskip3pt}@{\hskip3pt}c@{\hskip3pt}@{\hskip3pt}c@{\hskip3pt}c}
				\textbf{Method} & \textbf{12 bits} & \textbf{24 bits} & \textbf{36 bits} & \textbf{48 bits} \\
				\hline
				DSH &  $ {0.761 \pm 0.018}$ & $ {0.792 \pm 0.012}$ & $ {0.809 \pm 0.008}$ & $ {0.819 \pm 0.002}$ \\
				DPSH &  $ {0.767 \pm 0.023}$ & $ {0.773 \pm 0.005}$ & $ {0.774 \pm 0.008}$ & $ {0.759 \pm 0.007}$ \\
				QSMIH &  $ \mathbf{0.842 \pm 0.012}$ & $ \mathbf{0.857 \pm 0.004}$ & $ \mathbf{0.858 \pm 0.007}$ & $ \mathbf{0.861 \pm 0.004}$ \\
			\end{tabular}
		\end{center}
	\end{table}
	
	\subsection{Supervised Hashing Evaluation}

	The proposed method was also evaluated using the  CIFAR10~\cite{krizhevsky2009learning} and NUS-WIDE~\cite{nus-wide-civr09} datasets. For the CIFAR10, a DenseNet~\cite{huang2017densely}, pretrained on the same dataset, was employed, while for the NUS-WIDE dataset, a DenseNet pretrained on the ImageNet dataset was used~\cite{ILSVRC15}.

	\begin{table}

		\caption{CIFAR10 Evaluation  (the mAP for different hash code lengths is reported)}

		\label{table:cifar10}

		\begin{center}

			\begin{tabular}{l|ccccc}

				\textbf{Method} & \textbf{8 bits} & \textbf{12 bits} & \textbf{24 bits} & \textbf{36 bits} & \textbf{48 bits} \\
				\hline
				MIHash* & N/A & 0.929 & 0.933 & 0.938*** & 0.942\\
				\hline
				DSH** &  $ {0.936}$ & $ {0.958}$ & $ {0.967}$ & $ {0.970}$ & $ {0.970}$ \\
				DPSH** &  $ {0.776}$ & $ {0.933}$ & $ \mathbf{0.971}$ & $ \mathbf{0.971}$ & $ \mathbf{0.971}$ \\
				QSMIH &  $ \mathbf{0.962}$ & $ \mathbf{0.970}$ & $ \mathbf{0.971 }$ & $ \mathbf{0.971}$ & $ \mathbf{0.971}$ \\
			\end{tabular}
		\end{center}

		\scriptsize{*Results as reported in~\cite{Cakir_2017_ICCV} (using a slightly different setup), **Results using our implementation of DSH~\cite{liu2016deep} and DPSH~\cite{li2016feature}, ***Results for 32 bits (as reported in \cite{Cakir_2017_ICCV}).}

	\end{table}

	The evaluation results for the CIFAR10 dataset are reported in Table~\ref{table:cifar10}. The proposed method is also compared to the MIHash method ~\cite{Cakir_2017_ICCV}. The proposed method outperforms all the other techniques by a large margin for small code lengths, i.e., 8 and 12 bits. For larger hash codes, the proposed method performs equally well with the DSH and DPSH methods. However, the proposed method is capable of achieving almost the same performance as the DSH and DPSH methods using less than half the bits, highlighting the expressive power of the proposed technique. Also, to the best of our knowledge, this is the best result reported in the literature for the CIFAR10 dataset (regardless the employed hashing technique). The Hamming precision within radius 2 is shown in Figure~\ref{fig:hamming-cifar}. Again, the proposed method performs significantly better for smaller code lengths, i.e., 8 and 12 bits, while matching the precision of the DPSH method for hash codes larger than 24 bits.

	The proposed QSMIH method was also evaluated using the larger-scale NUS-WIDE dataset that contains 269,648 images that belong to 81 different concepts. Following~\cite{li2016feature}, we used the images that belong to the 21 most frequent concepts, i.e., 195,834 images. Note that an image might belong to more than one concept, i.e., fulfill multiple information needs. Furthermore, instead of using a subsample of the training set, we used the whole training set (193,734 randomly sampled images) to learn the hash codes (all the methods were used in a batch setting), and a test set of 2,100 randomly sampled queries was employed to evaluate the methods. Since there are many differences in the evaluation protocol used by different papers for this dataset, we compared the proposed method to the DSH and DPSH methods using the same network and evaluation setup. The evaluation results are shown in Table~\ref{table:nuswide}. Again, the proposed method outperforms the rest of the evaluated methods for any code length. The same behavior is also observed in the precision results illustrated in Fig.~\ref{fig:hamming-nuswide}. Note that a similar behavior for the DSH method, i.e., the precision is reduced as the code length increases after a certain point, is also reported by the authors of the DSH method~\cite{liu2016deep}.

	\begin{table}
		\caption{NUS-WIDE Evaluation (the mAP for different hash code lengths is reported)}
		
		\label{table:nuswide}
		\begin{center}
			\begin{tabular}{l|ccccc}
				\textbf{Method}  & \textbf{8 bits} & \textbf{12 bits} & \textbf{24 bits} & \textbf{36 bits} & \textbf{48 bits}\\
				\hline
				
				DSH &  $ {0.660 }$ & $ {0.659}$ & $ {0.671}$ & $ {0.689}$ & $ {0.694}$ \\
				
				DPSH &  $ {0.735}$ & $ {0.748}$ & $ {0.759}$ & $ {0.758}$ & $ {0.755}$ \\
				
				QSMIH &  $ \mathbf{0.746}$ & $\mathbf{0.753}$ & $ \mathbf{0.766}$ & $ \mathbf{0.764}$ & $ \mathbf{0.763}$ \\
				
			\end{tabular}

		\end{center}

	\end{table}

	\subsection{Retrieval of Unseen Information Needs}
	\label{section:eval-unseen}
	Finally, the proposed method was evaluated using the evaluation setup proposed in~\cite{sablayrolles2017should}, i.e., the 75\% of the classes were used to train the models and the rest 25\% were used to evaluate the models. The process was repeated 5 times using different class/information needs splits and the mean and standard deviation are reported. The evaluation results are shown in Table~\ref{table:cifar10-unseen}. The proposed method also employed 5 unsupervised information needs (discovered using the k-means algorithm on the 75\% of the training data). The weight of the unsupervised loss in the optimization was set to $\beta=0.5$. The proposed variant, denoted by ``QSMIH+U'' leads to significantly more regularized representations, that do not collapse outside the training domain, increasing the mAP for unseen classes from 0.689 to 0.795, demonstrating the flexibility of the proposed approach as well as its effectiveness in this setup.

	\begin{table}
		\caption{CIFAR10 Evaluation - Retrieval of Unseen Information Needs (the mAP for different hash code lengths is reported)}
		
		\label{table:cifar10-unseen}
		\begin{center}
			\begin{tabular}{l|ccccc}
				\textbf{Method}  &  \textbf{12 bits} & \textbf{24 bits} & \textbf{36 bits}\\
				\hline
				DSH &  $ {0.615 \pm 0.065}$ & $ {0.689 \pm 0.063}$ & $ {0.674 \pm 0.055}$ \\
				DPSH &  $ {0.568 \pm 0.082}$ & $ {0.635 \pm 0.078}$ & $ {0.658 \pm 0.048}$ \\
				QSMIH+U &  $\mathbf{0.691 \pm 0.093}$ & $\mathbf{0.795 \pm 0.061}$ & $\mathbf{0.682 \pm 0.144}$ \\
				
			\end{tabular}

		\end{center}

	\end{table}

	\section{Conclusions}

	\label{section:conclusions}

	A deep supervised hashing algorithm,  adapted to the needs of large-scale hashing, that optimizes the learned codes using an novel information-theoretic measure, the Quadratic Spherical Mutual Information, was proposed. The proposed method was evaluated using three different datasets and evaluation setups and compared to other state-of-the-art supervised hashing techniques. The proposed method outperformed all the other evaluated methods regardless the size of the used dataset and training setup, exhibiting a significantly more stable behavior than the rest of the evaluated methods. More specifically, when used with a randomly initialized network, the proposed QSMIH method managed to outperform the rest of the methods by a large margin. On the other hand, when combined with powerful pre-trained networks, again it yielded the best results regardless the length of the used hash code. Also, the proposed method provides theoretical justification for several existing deep supervised hashing techniques, while also paves the way for developing more advanced representation learning techniques for information retrieval using the proposed information-theoretic formulation, e.g., handling cross-modal retrieval tasks~\cite{xu2017learning}.

	\section*{Appendix A - Implementation Details}

	To simplify the implementation of the proposed method, we assume that all the information needs are equiprobable: $
	P(q) = \frac{1}{M}.$ Then, the information potentials $V^{cos}_{ALL}(\mathscr{Q}, \mathscr{Y})$ and $V^{cos}_{BTW}(\mathscr{Q}, \mathscr{Y})$ can be calculated as:
	\begin{equation}
	\begin{aligned}
		V^{cos}_{ALL} (\mathscr{Q}, \mathscr{Y}) = \frac{1}{N^2} \left( \sum_{k=1}^M \left(\frac{N_k}{N}\right)^2\right)  \sum_{i=1}^{N} \sum_{j=1}^{N} S_{cos}(\mathbf{y}_i, \mathbf{y}_j) \\
		= \frac{1}{N^2} \left( \sum_{k=1}^M \left(\frac{1}{M}\right)^2\right)  \sum_{i=1}^{N} \sum_{j=1}^{N} S_{cos}(\mathbf{y}_i, \mathbf{y}_j)
		= \frac{1}{N^2} \frac{1}{M}  \sum_{i=1}^{N} \sum_{j=1}^{N} S_{cos}(\mathbf{y}_i, \mathbf{y}_j),
	\end{aligned}		
	\end{equation}
	and
	\begin{equation}
	\begin{aligned}	
		V^{cos}_{BTW} (\mathscr{Q}, \mathscr{Y}) = \frac{1}{N^2} \sum_{k=1}^{M} \left( \left(\frac{N_k}{N}\right)  \sum_{i=1}^{N_k} \sum_{j=1}^{N}  S_{cos}(\mathbf{y}^{(k)}_i, \mathbf{y}_j) \right)\\
		= \frac{1}{N^2} \frac{1}{M}  \sum_{k=1}^{M}  \sum_{i=1}^{N_k} \sum_{j=1}^{N}  S_{cos}(\mathbf{y}^{(k)}_i, \mathbf{y}_j)\\
		= \frac{1}{N^2} \frac{1}{M}  \sum_{i=1}^{N} \sum_{j=1}^{N} S_{cos}(\mathbf{y}_i, \mathbf{y}_j) = V^{cos}_{ALL} (\mathscr{Q}, \mathscr{Y}),
	\end{aligned}				
	\end{equation}
	where we assumed that $N_k = \frac{N}{M}$, since $P(q) = \frac{1}{M}$. Also, the $V^{cos}_{IN}(\mathscr{Q}, \mathscr{Y})$ information potential can be expressed using the indicator matrix $\mathbf{\Delta}$:
	\begin{equation}
		V^{cos}_{IN} (\mathscr{Q}, \mathscr{Y}) 
		= \frac{1}{N^2} \sum_{i=1}^{N}  \sum_{j=1}^{N_k} [\mathbf{\Delta}]_{ij}S_{cos}(\mathbf{y}_i, \mathbf{y}_j).
	\end{equation}
	Therefore, the QSMI can be simplified as:
	\begin{equation}
		\label{eq:3}
		I^{cos}_T (\mathscr{Q}, \mathscr{Y}) = V^{cos}_{IN} (\mathscr{Q}, \mathscr{Y})  - V^{cos}_{BTW} (\mathscr{Q}, \mathscr{Y})\\
		=\frac{1}{N^2} \sum_{i=1}^{N}  \sum_{j=1}^{N}\left([\mathbf{\Delta}]_{ij}S_{cos}(\mathbf{y}_i, \mathbf{y}_j) - \frac{1}{M}S_{cos}(\mathbf{y}_i, \mathbf{y}_j)\right)
	\end{equation}
	
	Finally, using the proposed clamping method, the final loss function is obtained as:
	\begin{equation}
		\label{eq:4}
		\mathcal{L}_{QSMI} =\frac{1}{N^2} \sum_{i=1}^{N}  \sum_{j=1}^{N}\left([\mathbf{\Delta}]_{ij} \left(S_{cos}(\mathbf{y}_i, \mathbf{y}_j) - 1\right)^2 + \frac{1}{M} \left(S_{cos}(\mathbf{y}_i, \mathbf{y}_j)\right)^2 \right)
	\end{equation}
	since we aim to maximize (\ref{eq:3}). Also, note that (\ref{eq:4}) can be equivalently expressed as:
	\begin{equation}
		\label{eq:qsmi-obj}
		\mathcal{L}_{QSMI} = \frac{1}{N^2} \mathbf{1}^T_N \left(\mathbf{\Delta}\odot (\mathbf{S}-1)\odot(\mathbf{S}-1) + \frac{1}{M}(\mathbf{S}\odot\mathbf{S})\right) \mathbf{1}_N
	\end{equation}
	allowing for efficiently implementing the proposed method.

	To implement the gradient descent algorithm, the derivative $\frac{\partial \mathcal{L}_{QSMI}}{\partial \mathbf{W}}$, where $\mathbf{W}$ are the parameters of the employed neural network, must be calculated. This derivative is calculated as:
	\begin{equation}
		\frac{\partial \mathcal{L}_{QSMI}}{\partial \mathbf{W}} = \sum_{i=1}^{N} \left( \frac{\partial \mathcal{L}_{QSMI}}{\partial \mathbf{y}_i} \right)^T \frac{\partial \mathbf{y}_i}{\partial \mathbf{W}},
	\end{equation}
	where $\mathbf{y}_i$ is the output of the used neural network for the $i$-th document. The derivative $ \frac{\partial \mathbf{y}_i}{\partial \mathbf{W}}$ depends on the employed architecture, while the derivative of the proposed loss with respect to the hash code $\frac{\partial \mathcal{L}_{QSMI}}{\partial \mathbf{y}_i}$ can be calculated as:
	
	\begin{equation}
		\frac{\partial \mathcal{L}_{QSMI}}{\partial \mathbf{y}_i} =   \frac{1}{N^2}  \sum_{j=1, i\neq j}^{N} \left([\mathbf{\Delta}]_{ij} \frac{\partial \left(S_{cos}(\mathbf{y}_i, \mathbf{y}_j) - 1\right)^2}{\partial \mathbf{y}_i} - \frac{1}{M} \frac{\partial\left(S_{cos}(\mathbf{y}_i, \mathbf{y}_j)\right)^2}{\partial \mathbf{y}_i} \right),
	\end{equation}
	where
	\begin{equation}
		\label{eq:1}
		\frac{\partial \left(S_{cos}(\mathbf{y}_i, \mathbf{y}_j) - 1\right)^2}{\partial \mathbf{y}_i} = 2 \left(S_{cos}(\mathbf{y}_i, \mathbf{y}_j) - 1\right) \frac{\partial S_{cos}(\mathbf{y}_i, \mathbf{y}_j)}{\partial \mathbf{y}_i},
	\end{equation}
	and
	\begin{equation}
		\label{eq:2}
		\frac{\partial\left(S_{cos}(\mathbf{y}_i, \mathbf{y}_j)\right)^2}{\partial \mathbf{y}_i}= 2 S_{cos}(\mathbf{y}_i, \mathbf{y}_j) \frac{\partial S_{cos}(\mathbf{y}_i, \mathbf{y}_j)}{\partial \mathbf{y}_i}.
	\end{equation}
	
	Finally, the derivative of the cosine similarity, needed for calculating (\ref{eq:1}) and~(\ref{eq:2}), can be computed as:
	\begin{equation}
		\frac{\partial S_{cos}(\mathbf{y}_i, \mathbf{y}_j)}{\partial [\mathbf{y}_i]_l} = \frac{1}{2}\left(
		\frac{[\mathbf{y}_j]_l}{\Vert \mathbf{y}_i \Vert_2 \Vert \mathbf{y}_j \Vert_2}
		- \frac{\mathbf{y}_i^T \mathbf{y}_j}{\Vert \mathbf{y}_i \Vert_2 \Vert \mathbf{y}_j \Vert_2}
		\frac{[\mathbf{y}_i]_l}{\Vert \mathbf{y}_i \Vert_2^2}
		\right).
	\end{equation}
	
	The loss and the corresponding derivatives can be similarly calculated when the information needs are not equiprobable.

	\section*{Appendix B - Hyper-parameters and Network Architectures}

	The selected hyper-parameters are shown in Table~\ref{table:parameters}. We selected the best parameters for the two other evaluated methods, i.e., DSH and DPSH, by performing line search for each parameter. The Adam optimizer~\cite{adam}, with the default hyper-parameters, was used for the optimization. The experiments were repeated 5 times and the mean value of each of the evaluated metrics is reported, except otherwise stated.
	
	\begin{table}
		\caption{Parameters used for the conducted experiments}
		\label{table:parameters}

		\begin{center}
			\begin{tabular}{l|c|ccc}
				\textbf{Parameter} & \textbf{Method} & \textbf{F. MNIST} & \textbf{CIFAR10} & \textbf{NUS-WIDE}\\
				\hline
				Learning rate & all & 0.001 & 0.001 & 0.001 \\
				Batch size & all  & 128 & 128 & 128 \\
				Epochs &  all  & 50 & 5 & 50 \\
				$\alpha$ & DSH & $10^{-5}$  & $10^{-5}$ &  $10^{-5}$\\
				$\alpha$ & QSMIH & $10^{-2}$  & $10^{-2}$ &  $10^{-1}$\\	
				$\eta_{DPSH}$~\cite{li2016feature} & DPSH & 5 & $3$ &  $5$\\		
				\multicolumn{2}{l}{}& (3 for 36-48 bits) \\
			\end{tabular}
		\end{center}
	\end{table}
	
	For the experiments conducted on the Fashion MNIST dataset a relatively simple Convolutional Neural Network (CNN) architecture was employed, as shown in Table~\ref{table:architecture}. The network was initialized using the default PyTorch initialization scheme~\cite{paszke2017automatic}, and it was trained from scratch for all the conducted experiments.

	For the CIFAR10 dataset, a DenseNet-BC-190 (growth rate 40 and compression rate 2)~\cite{huang2017densely}, that was pretrained on the CIFAR dataset, was used. For the NUS-WIDE dataset, a DenseNet-201 (growth rate 32 and compression rate 2), that was pretrained on the Imagenet dataset~\cite{ILSVRC15}, was also employed. The feature representation was extracted from the last average pooling layers of the networks. Then, two fully connected layers were used: one with $N_H$ neurons and rectifier activation functions~\cite{nair2010rectified}, and one with as many neurons as the desired code length (no activation function was used for the output layer). The size of hidden layer was set to $N_H=64$ for the CIFAR10 dataset and to $N_H=2048$ for the NUS-WIDE dataset. To speedup the training process, we back-propagated the gradients only to the last two layers of the network, which were trained to perform supervised hashing.

	\begin{table}
		\caption{Network architecture used for the Fashion MNIST dataset}
		\label{table:architecture}
		\begin{center}
			\begin{tabular}{l |ccc}
				\textbf{Layer} & \textbf{Kernel Size} & \textbf{Filters / Neurons} & \textbf{Activation} \\
				Convolution & $5 \times 5$ & 32 & ReLU~\cite{nair2010rectified} \\
				Max Pooling & $2 \times 2$ & - & - \\
				\hline
				Convolution & $5 \times 5$ & 64 &  ReLU~\cite{nair2010rectified} \\
				Max Pooling & $2 \times 2$ & - & - \\	
				\hline
				Dense & - & \# bits & -\\
			\end{tabular}
			
		\end{center}

	\end{table}

\bibliographystyle{plain}
\bibliography{refs}
	
\end{document}